\documentclass[10pt,sigconf,letterpaper,nonacm]{acmart}
\usepackage{algorithmic}
\usepackage{graphicx}
\usepackage{textcomp}
\usepackage{enumitem}
\usepackage{multirow}
\usepackage{balance}
\usepackage{lscape}
\usepackage{subcaption}
\usepackage{makecell}
\usepackage{tabularray}

\usepackage{algorithm}
\usepackage{algorithmic}

\usepackage{threeparttable}

\AtBeginDocument{%
  }

\usepackage{xspace}

\begin{document}

%
\title{Towards a graph-based foundation model for network traffic analysis}

\author{Louis Van Langendonck}
\email{louis.van.langendock@estudiantat.upc.edu}
\affiliation{%
  \institution{Universitat Polit{\'e}cnica de Catalunya}
  \city{Barcelona}
  \country{Spain}
}

\author{Ismael Castell-Uroz}
\email{ismael.castell@upc.edu}
\affiliation{%
  \institution{Universitat Polit{\'e}cnica de Catalunya}
  \city{Barcelona}
  \country{Spain}
}

\author{Pere Barlet-Ros}
\email{pere.barlet@upc.edu}
\affiliation{%
  \institution{Universitat Polit{\'e}cnica de Catalunya}
  \city{Barcelona}
  \country{Spain}
}

\renewcommand{\shortauthors}{Van Langendonck et al.}

\begin{abstract}
Foundation models have shown great promise in various fields of study. A potential application of such models is in computer network traffic analysis, where these models can grasp the complexities of network traffic dynamics and adapt to any specific task or network environment with minimal fine-tuning. Previous approaches have used tokenized hex-level packet data and the model architecture of large language transformer models. We propose a new, efficient graph-based alternative at the flow-level. Our approach represents network traffic as a dynamic spatio-temporal graph, employing a self-supervised link prediction pretraining task to capture the spatial and temporal dynamics in this network graph framework. To evaluate the effectiveness of our approach, we conduct a few-shot learning experiment for three distinct downstream network tasks: intrusion detection, traffic classification, and botnet classification. Models finetuned from our pretrained base achieve an average performance increase of 6.87\% over training from scratch, demonstrating their ability to effectively learn general network traffic dynamics during pretraining. This success suggests the potential for a large-scale version to serve as an operational foundational model.

\end{abstract}




\maketitle

\section{Introduction}
Recent proposals considering large-scale pretraining for task-agnostic foundation models have proven transformative in the academic world and beyond. A variant of such model applied to computer network traffic could provide similar impact, though a successful implementation has yet to be put forth. Such a network traffic foundation model should understand complex network traffic dynamics and thus effortlessly specialize to any downstream task and any network setting with minimal fine-tuning.

The key components of any foundation model are: a consistent data representation and model architecture that efficiently captures the dynamics of the application field, an informative self-supervised pretraining task enabling general knowledge acquisition, and using large amounts of data to scale up the pretraining step. 
Previous approaches to developing a network traffic foundation model employ large language transformer models to operate on tokenized hex-level packet data. In this paper we put forth a new, efficient graph-based alternative for the first two key elements (data representation \& model architecture and pretraining task), hence laying the foundation for an eventual large-scale implementation. 

We argue in favour of a graph-based approach for the following reasons: $(i)$ Graphs are well-suited to represent the structural characteristics of network traffic, capturing the relationships and interactions between different nodes and connections more accurately than sequential tokenization \cite{egraphsage, anomale, gnn_for_encrypted}. $(ii)$ Working at the flow-level is more efficient and informative because it aggregates data, providing a clearer and more meaningful representation of network traffic compared to the limited insights available from encrypted packet-level data \cite{enisa, gnn_for_encrypted}. $(iii)$ The complex inter-dependencies and long-term relationships in packet-level data requires extremely large and resource-intensive models for effective processing. In contrast, GNNs are more efficient to train and deploy due to geometrical reuse of parameters, with our model being on average only 1\% the size of other network traffic foundation model proposals \cite{graphsage, gnn_power, bert, et_bert, lens, netfound}.

To show that the method we put forth indeed effectively learns general network traffic behaviour through pretraining, we set up a few-shot learning experiment for three distinct down-stream network tasks: intrusion detection, traffic classification and botnet classification. For each task, with limited data and training epochs, one GNN model is trained from scratch while another GNN model is finetuned from our pretrained base, which is trained on two small unlabeled datasets considering a completely different network setting. The results show that the models using our pretrained base show an average performance increase of 6.87 \% compared to training from scratch. Although only a small amount of pretraining data is used, completely independent from the data used for fine-tuning, we conclude that knowledge is effectively transferred from pretrained base to finetuned models in unseen down-stream tasks, underlining the potential for a large-scale foundation model.

The rest of the paper is organized as follows: Section \ref{related_work} reviews the background and the related work. Section \ref{proposal} describes the characteristics of our proposed model while Section \ref{evaluation} presents the few-shot experiment and its results. Section \ref{conclusions} concludes the paper.

\section{Related Work}
\label{related_work}

The few existing proposals for a network traffic foundation model typically consider a large language transformer model operating on tokens of raw, encrypted packet-level data in hex-format \cite{lens, netfound, et_bert}. They rely on the observation that encrypted traffic is not perfectly random and both implicit and structural patterns exist \cite{et_bert, encrypted_not_random}. 
 \textit{Lens} \cite{lens}, proposed by Wang, et al., employs the T5 model, a popular text-to-text encoder-decoder transformer model. Three self-supervised pretraining tasks are proposed: masked span prediction, packet order prediction and homologous traffic prediction.  They show that their finetuned models often perform better on several down-stream tasks than other packet-level models. \textit{ET-BERT} \cite{et_bert}, put forth by Lin, et al., instead use as model architecture the well know language foundation model \textit{BERT} (bidirectional Encoder Representations from Transformers). They consider two pretraining tasks: token prediction and origin prediction. Their finetuned models generally outperform the state-of-the-art in different encrypted classification tasks. \textit{netFound} \cite{netfound}, proposed by Guthula, et al., increases complexity compared to the aforementioned works by accounting for mixed granularity levels (packets, bursts, flows) in the model architecture. Additionally to tokenization, several statistical features are conjointly fed to the BERT-inspired model. Extra aggregation layers enable final representations at different granularity levels. The only pretraining task they consider is token prediction. They outperform most other classification models operating on packet-level data. 

Although these solutions are promising, we claim that they fall short in several regards: $(i)$ They fail to capture the structural characteristics of network traffic, as sequential tokenization does not accurately model relationships and dependencies between network elements. $(ii)$ Working at the packet level is less efficient and informative compared to strictly working at the flow level, as encrypted packet data often lacks context and useful features. $(iii)$ The complex inter-dependencies and long-term relationships in packet-level data require extremely large and resource-intensive models, making them impractical for effective processing.

\begin{figure*}
    \centering
    \includegraphics[width=\textwidth]{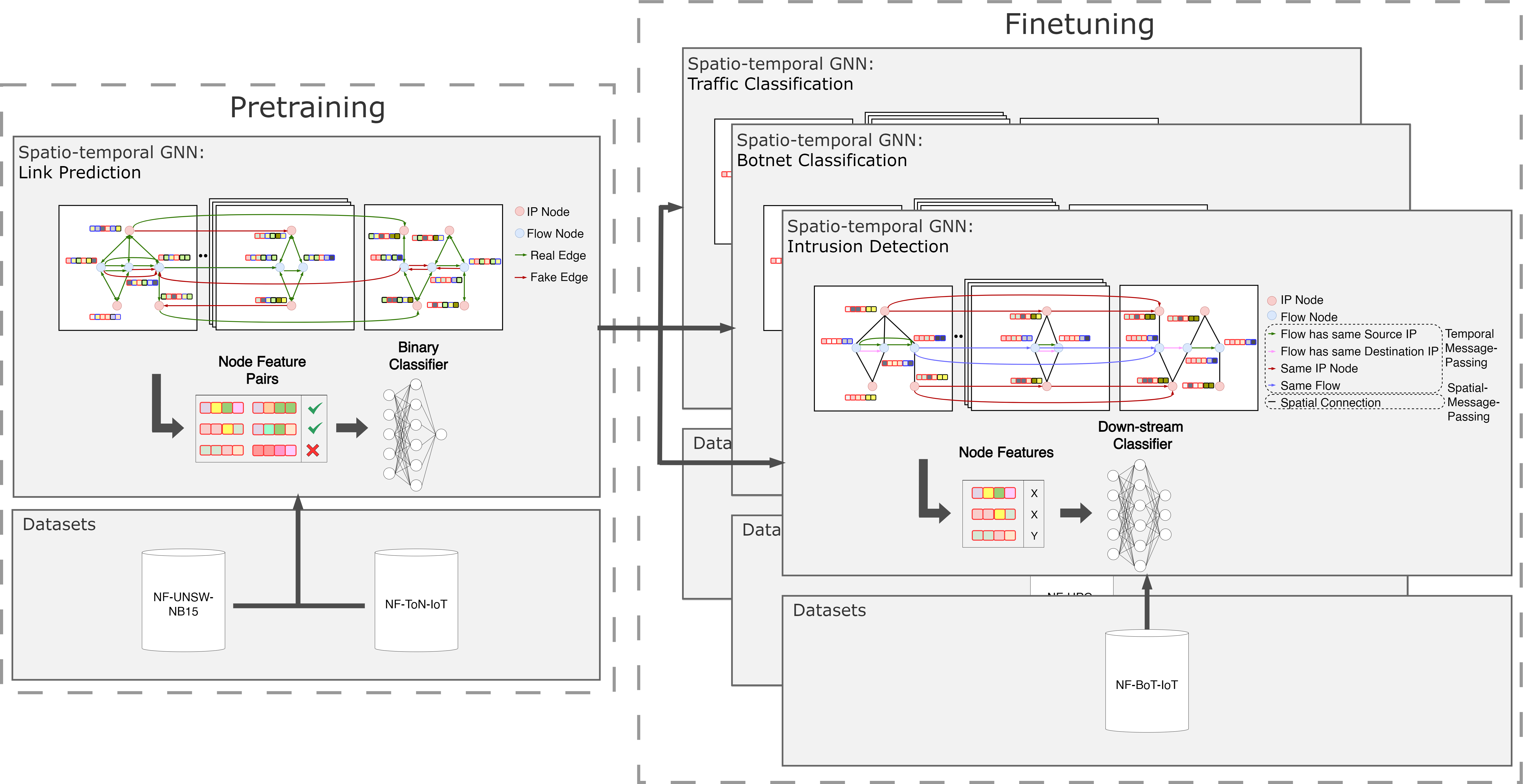}
    \caption{Overview of proposed model and workflow. \textmd{A graph neural network model is pretrained using an unsupervised link-prediction task. The resulting model is used for finetuning to different downstream tasks. The Graph Neural Network model incorporates spatio-temporal information and uses both structural and feature-based information to update node representations.}}
    \label{fig:workflow}
\end{figure*}

\section{Proposal}
\label{proposal}
In this section we introduce the architecture and training routine of the proposed computer network traffic foundation model. Figure \ref{fig:workflow} provides a high-level overview of the proposed workflow. During unsupervised pretraining, we use unlabeled, "cheap" network data to teach a graph-based model the essential behaviour characterizing network traffic. Continuing from the pretrained model weights, the finetuning step consists of further supervised training on different down-stream network analysis tasks, using limited labeled data. We use different datasets for pretraining and for finetuning, thus avoiding data leakage and allowing for a fair comparison to down-stream models trained from scratch. Note that each down-stream model is finetuned independently though all starting from the same pretrained base. 

First we summarize the chosen network graph representation and corresponding graph neural network architecture. Afterwards we go in to detail about the proposed pretraining task. Finally we describe the finetuning process.

\subsection{Model Architecture} \label{model_architecture}

A schematic representation of the network graph and corresponding graph neural network model is displayed in the finetuning section of the general proposal overview depicted in Figure \ref{fig:workflow}. 


We opt for the line-graph representation of the network traffic. This entails that for every flow, we represent both the flow itself and the source and destination IP address as a node. Both source and destination IP are connected to the flow node using bidirectional spatial edges. The initial features of the flow node are the preprocessed flow features while the IP nodes are initialized with dummy one-vectors. 

In order to include a temporal component, we build several small graphs spanning a few second of network traffic and add temporal edges between these of reoccuring nodes. Within each window, temporal edges are included for indicating the order of flow nodes from the same source or destination IP. Further details can be found in our previous work using the same graph-building method \cite{ppt-gnn}.


We opt for a spatio-temporal graph neural network model to capture the dynamic nature of the proposed network graph. Generally speaking, this architecture can be considered a standard heterogeneous, convolutional graph neural network, implemented using the GraphSAGE \cite{graphsage} framework, that is extended to effectively handle the temporal component. Again, further details can be found in our previous work employing the same model \cite{ppt-gnn}.

However, it is important to note that alternative graph representations and model architectures could also yield good results and should be considered in future work. We made the considered choice of this spatio-temporal approach because of the high expressiveness of the model.

Note that most graph neural networks are relatively compact compared to transformer-based models due to the reuse of parameters across the graph. Our proposed model contains about $732.000$ parameters, which on average is about 1\% of the size of the models of other proposed network traffic foundation models (\textit{ET-BERT}: $\sim 110$ Million \cite{bert, et_bert}, \textit{netFound}: $\sim 220$ Million \cite{netfound}, \textit{Lens}: $ \sim 60 - 770$ Million \cite{lens, t5})

\subsection{Pretraining}
Effective pretraining is crucial in developing a successful foundation model. The pretraining task should be self-supervised, only make use of standardized, unlabeled network data to learn network traffic dynamics. Several graph-based, self-supervised pretraining methods have recently been proposed ranging from context prediction to node-level feature prediction \cite{always_be_pretraining, graph_foundation_model, learning_to_pretrain, pretraining_strategies}. As a proof of concept, we opt for the simple task of link prediction. However, note that a large-scale implementation, just like in the large language setting, would benefit from multiple pretraining tasks combined in a single loss function \cite{bert, gpt3}. Our proposed link-prediction task considers the following steps: $(i)$ \textbf{Negative Sampling:} For any edge type, fake edges are randomly added. In each graph, we choose to have an equal amount of fake edges as real edges $(ii)$ \textbf{Link Prediction:} The GNN model described in Section \ref{model_architecture} is given the binary classification task to classify every edge as real or not. This teaches the model the spatial and temporal dynamics of the network graph, learning the typical structural and feature-based relations between different node types and across different edge types. A schematic representation of the link-prediction task is displayed in the pretraining section of the general proposal overview depicted in Figure \ref{fig:workflow}.


\subsection{Finetuning}

The final graph representation obtained by pretrained model can now be used for efficient finetuning to a downstream task in network traffic analysis. First a graph neural network model is initialized with the weights from the pretrained model. Some simple fully-connected neural network layers are connected to the final pretrained representations of the flow nodes. Then, the entire model is finetuned in a supervised manner using labeled data (of e.g. a network intrusion detection task). The acquired understanding of network traffic dynamics during pretraining allows for limiting the amount of labeled data and training epochs, generally making the finetuning process an inexpensive process.

In this work we consider three network traffic analysis tasks: network classification, network intrusion detection and botnet classification. Note that although each of these tasks consider flow classification, the pretrained base can just as well be used for graph-level or edge-level classification or regression tasks. 

\section{Evaluation}
\label{evaluation}

\subsection{Downstream tasks}
\label{eval:datasets}
As already explained, our objective is to set up a few-shot experiment, comparing generalization capabilities training from scratch to fine-tuning from our pretrained model. Since the pretraining and fine-tuning datasets are strictly separated and consider different network settings, the pretrained model needs to learn general network dynamics in order to improve performance in fine-tuning compared to training from scratch. Therefore, the experiment is well-suited to assess the potential of the pre-trained, spatio-temporal graph neural networks as network traffic foundation model.

The three down-stream tasks that we will use for the few-shot study are:
\begin{itemize}
    \item \textbf{Intrusion Detection Systems}: An IDS is a system trained to detect different network attacks such as Denial of Services attacks or Network Scans. These systems represent a key element for network security experts that use them for early-detection of possible service outages or system intrusion. In particular, we use three different IDS datasets publicly available: \textbf{UNSW-NB15~\cite{unswnb15}}, \textbf{ToN-IoT~\cite{toniot}} and \textbf{BoT-IoT~\cite{botiot}}. The first two of them are used during the pretraining phase, while the last one is used as one of the downstream tasks to study. The pretraining is done on the unlabeled version of the datasets.
    \item \textbf{Traffic Classification}: In computer networks, traffic classification refers to the identification of each network packet or flow with the application that generates it. While network attacks are somewhat limited in number, a network trace may contain hundreds or even thousands of different applications. In this work we use the \textbf{UPC-COMNET14~\cite{bujlow201575}} private dataset exclusively for the finetuning and downstream task of traffic classification problem.
    \item \textbf{Botnet Detection}: Botnets are group of infected machines controlled by a command center computer which can remotely execute tasks inside them. Usually, an attacker uses the command center to control the remote computers and perform network attacks to specific targets. The dataset used to study botnet detection as a downstream task is the publicly available \textbf{CTU13~\cite{garcia2014empirical}} dataset.
\end{itemize}

\begin{table}[]
    \centering
    \caption{Dataset/task details}
    \label{tab:datasets}
    \resizebox{0.475\textwidth}{!}{%
    \begin{tabular}{lccccc}
    \hline
    \textbf{Dataset} & \textbf{Size (raw)} & \textbf{\#Flows} & \textbf{\#Labels} & \textbf{Year} & \textbf{Task}                                                    \\ \hline
    CTU13     & 71.7GB & 262K & 7  & 2013 & \begin{tabular}[c]{@{}c@{}}Intrusion\\ Detection\end{tabular} \\
    UPC-COMNET14     & 54.2GB              & 882K             & 77                    & 2014          & \begin{tabular}[c]{@{}c@{}}Traffic\\ Classification\end{tabular} \\
    UNSW-NB15 & 99.1GB & 2M   & 10 & 2015 & Pretraining                                                   \\
    BoT-IoT   & 69.3GB & 72M  & 10 & 2018 & \begin{tabular}[c]{@{}c@{}}Botnet\\ Detection\end{tabular}    \\
    ToN-IoT   & 67.6GB & 30M  & 10 & 2019 & Pretraining                                                   \\ \hline
    \end{tabular}%
    }
\end{table}

All the datasets used are Netflow variants of the original datasets generated directly from network capture pcap files. The features collected are the same features of bidirectional flows explained in our previous work available at~\cite{ppt-gnn}. Table~\ref{tab:datasets} shows a summary of the characteristics of each dataset.

\begin{figure}
    \centering
    \includegraphics[width=\linewidth]{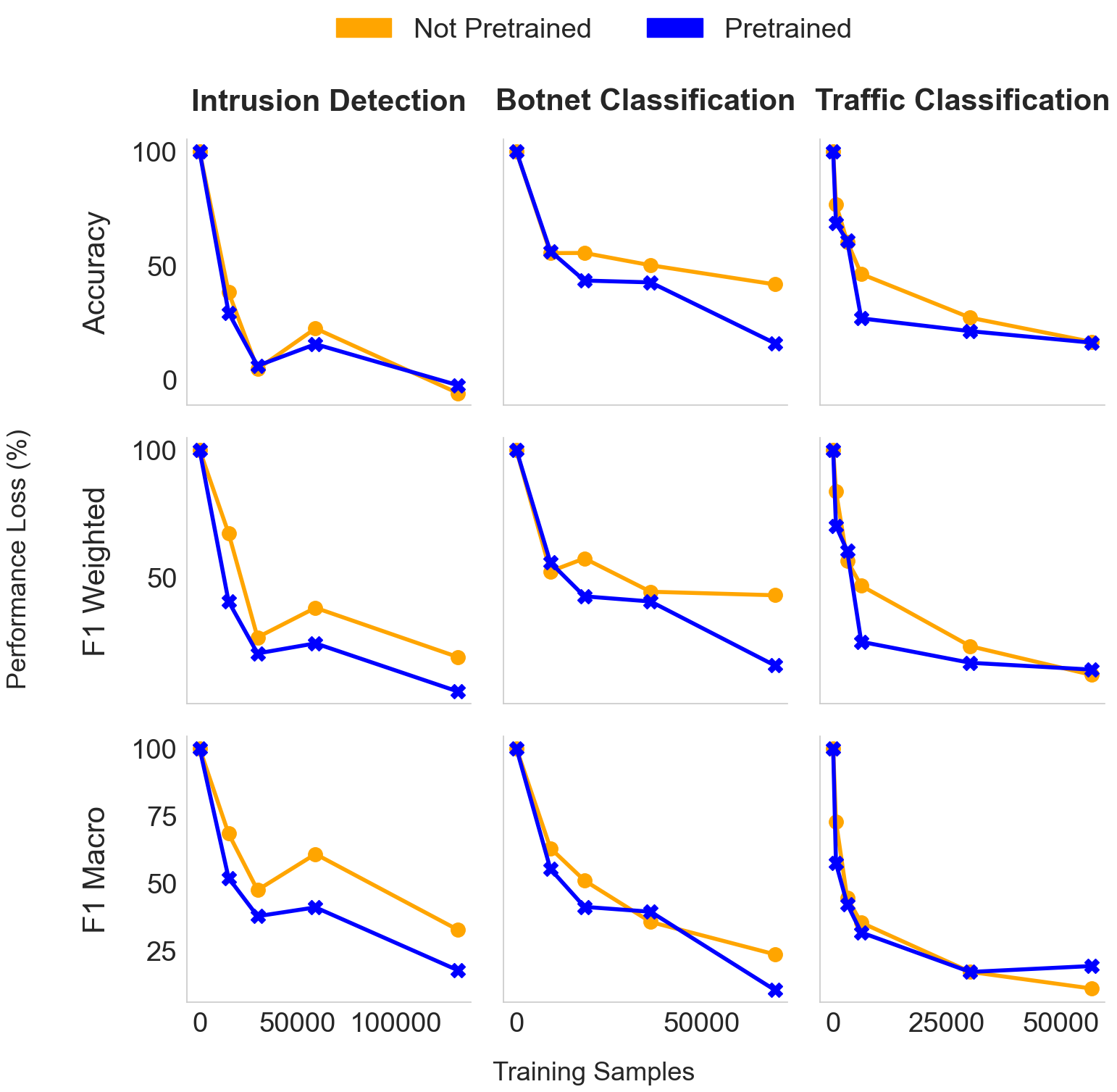}
    \caption{Few shot learning experiment. \textmd{For each of the downstream tasks (columns), one model is trained from scratch (orange) and one is finetuned from the pretrained model (blue). Each data point in the graph represents a different model. Training epochs are limited to 50. The results are expressed in percentage loss of optimal model performance of that metric (row) on the test set.}}
    \label{fig:few_shot}
\end{figure}

\subsection{Few-shot Experiment}

In the few-shot setting, we limit the amount of epochs used for training to 50. We slowly increase the amount of samples available for training, each time training two models: one trained from scratch and one finetuned from the pretrained base. In each training routine, we select the best model based on the multiclass macro f1 score, since it is the most challenging metric to optimize, taking into account both precision and recall as well as considering each class equally (including minority classes). Then each model is evaluated on the same dataset-specific test set, using three common multiclass classification metrics: accuracy, weighted f1 score and macro f1 score. 

Finally, serving as reference performance, for each dataset, a single model is trained using all data available for 200 epochs. The final metric used to evaluate performance is, for each of the three metrics, the percentual performance loss with respect to this reference performance. The results of this experiment are displayed in Figure \ref{fig:few_shot}. 

\begin{figure}
    \centering
    \includegraphics[width=\linewidth]{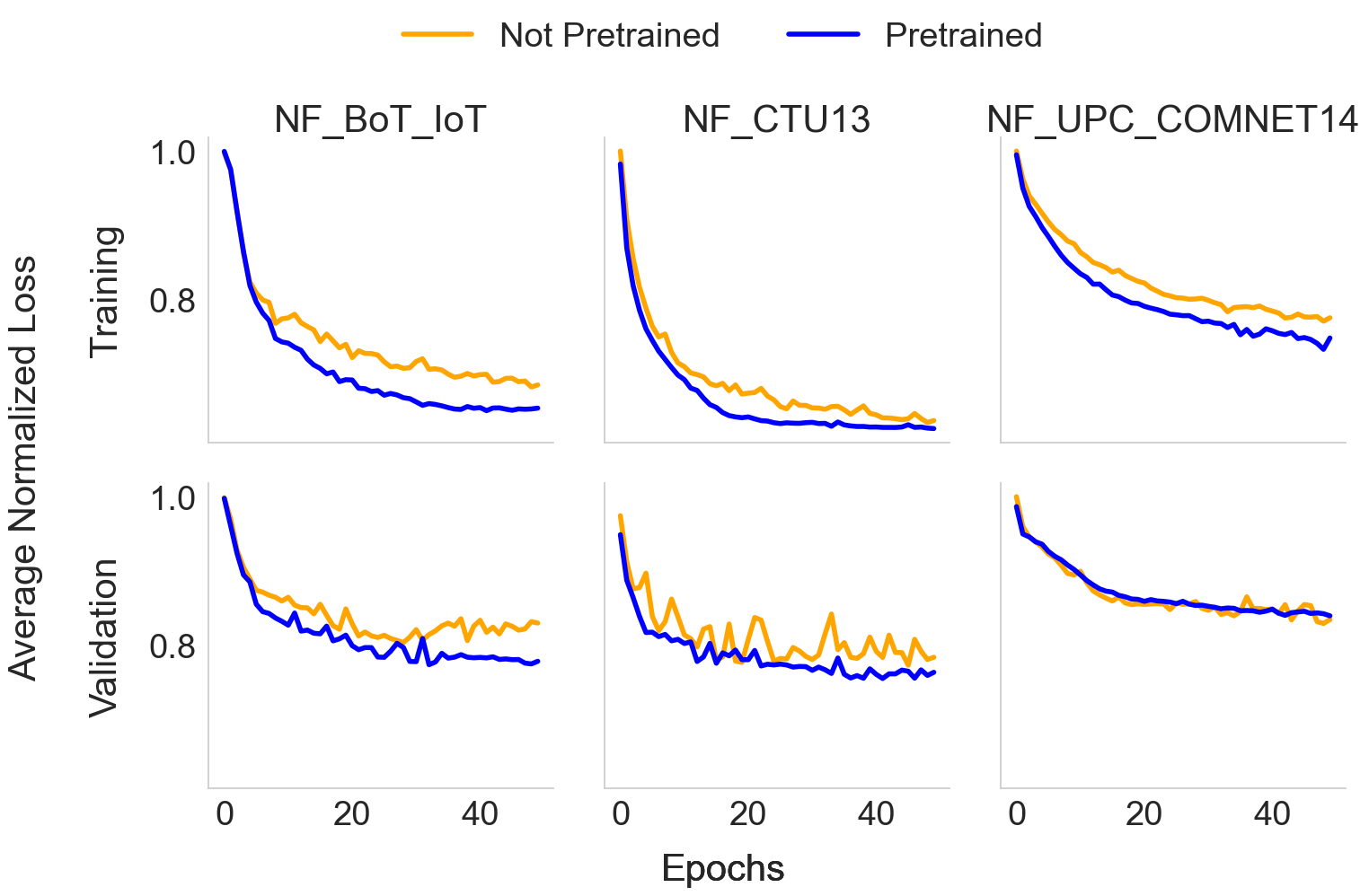}
    \caption{Average normalized loss curves. \textmd{For each of the downstream tasks (columns), the training and validation curves of the training routines of the few-shot experiment are normalized and then averaged at each epoch. This allows for a direct comparison of convergence speed and generalization capabilities between the finetuned models and the ones trained from scratch.}}
    \label{fig:loss_curves_norm_avg}
\end{figure}

To compare generalization speed between training from scratch and finetuning within each training cycle, we look at the training and validation loss curves for each of the models trained in the few-shot experiment. For direct comparison, we proceed to normalize the train and validation curves of each experiment and then average those results per epoch, per dataset and per training strategy (pretraining or from scratch). The resulting curves are plotted in Figure \ref{fig:loss_curves_norm_avg}.

The results of the few-shot experiment clearly show that pretraining the spatio-temporal graph neural network increases the model's generalization speed and performance in different down-stream tasks. For each of the datasets, for at least two of the metrics, the pretrained model approaches optimal performance sooner than without pretraining. The average difference across all datasets and metrics is 6.87 \%. The loss curves display similar tendencies: the training curves of the pretrained models minimizes faster and more profoundly for all three datasets. The validation curves show strong generalization in the pretrained case for both the \textit{NF-BoT-IoT} and \textit{NF-CTU13} datasets though not seeing any significant difference between curves in the \textit{NF-UPC-COMNET14} case. We believe this is due to this finetuning task of common network classification semantically being the furthest removed from the cybersecurity context of our pretraining dataset. This suggests adding more volume and variety to the pretraining dataset used in future work.

\section{Conclusions}
\label{conclusions}

This work presented a novel, graph-based approach to building network traffic foundation models. In contrast to previous works, our proposed method effectively incorporates the structural component of network traffic as well as reducing model size by about 99 \%. Via a few-shot experiment we show that by pretraining on a self-supervised link-prediction task using a small, out-of-context data sample, our proposed model successfully learns general network traffic dynamics, transferable to any down-stream networking task. Though significant up-scaling is required, the results offer a promising direction for operational network traffic foundation models. Looking ahead, future work will focus on increasing model complexity by including an attention-mechanism in message aggregation, increasing pretraining scale by adding several new unlabeled datasets and increasing pretraining complexity by including and combining multiple pretraining tasks.


\section*{Acknowledgements}
This work was supported by the CHISTERA grant CHIST-ERA-22-SPiDDS-02 corresponding to the GRAPHS4SEC project (reference nº PCI2023-145974-2) funded by the Agencia Estatal de Investigación through the PCI 2023 call.
This work is also supported by the Catalan Institution for Research and Advanced Studies (ICREA Academia).
\balance
\bibliographystyle{ACM-Reference-Format}
\bibliography{references}



\end{document}